\def\year{2020}\relax
\documentclass[letterpaper]{article}
\usepackage{aaai20}
\usepackage{times}
\usepackage{helvet}
\usepackage{courier}
\usepackage[hyphens]{url}
\usepackage{graphicx}
\usepackage{graphicx}
\usepackage{epsfig}
\usepackage{graphicx}
\usepackage{amsmath}
\usepackage{amssymb}
\usepackage{subfig}
\usepackage[utf8]{inputenc} 
\usepackage[T1]{fontenc}    
\usepackage{url}            
\usepackage{booktabs}       
\usepackage{amsfonts}       
\usepackage{nicefrac}       
\usepackage{microtype}      
\usepackage{graphicx}
\usepackage{algorithm}
\usepackage{algorithmic}
\usepackage{makecell}
\usepackage{mathtools}
\usepackage{multirow}
\usepackage{multicol}
\usepackage{color}

\newcommand{\etc}{\textit{etc.}}
\newcommand{\eg}{\textit{e.g.}}
\newcommand{\etal}{\textit{et al.}}
\newcommand{\ie}{\textit{i'e}}
\usepackage{amssymb}

\title{Sampling Strategies for GAN Synthetic Data }
\author{Binod Bhattarai, Seungryul Baek, Rumeysa Bodur, Tae-Kyun Kim\\ 
Imperial College London\\
\{b.bhattarai, s.baek15, r.bodur18, tk.kim\}@imperial.ac.uk
}

\begin{document}
\maketitle

\begin{abstract}
Generative Adversarial Networks (GANs) have been used widely to generate large volumes of synthetic data. This data is being utilized for augmenting with real examples in order to train deep Convolutional Neural Networks (CNNs). Studies have shown that the generated examples lack sufficient realism to train deep CNNs and are poor in diversity. Unlike previous studies of randomly augmenting the synthetic data with real data, we present our simple, effective and easy to implement synthetic data sampling methods to train deep CNNs more efficiently and accurately. To this end, we propose to maximally utilize the parameters learned during training of the GAN itself. These include discriminator's realism confidence score and the confidence on the target label of the synthetic data. 
In addition to this, we explore reinforcement learning (RL) to automatically search a subset of meaningful synthetic examples from a large pool of GAN synthetic data. 
We evaluate our method on two challenging face attribute classification data sets viz. AffectNet and CelebA. Our extensive experiments clearly demonstrate the need of sampling synthetic data before augmentation, which also improves the performance of one of the state-of-the-art deep CNNs 
\textit{in vitro}.
\end{abstract}
\section{Introduction}
\label{intro}

\begin{figure}
    \centering
    \includegraphics[width=1.0\linewidth]{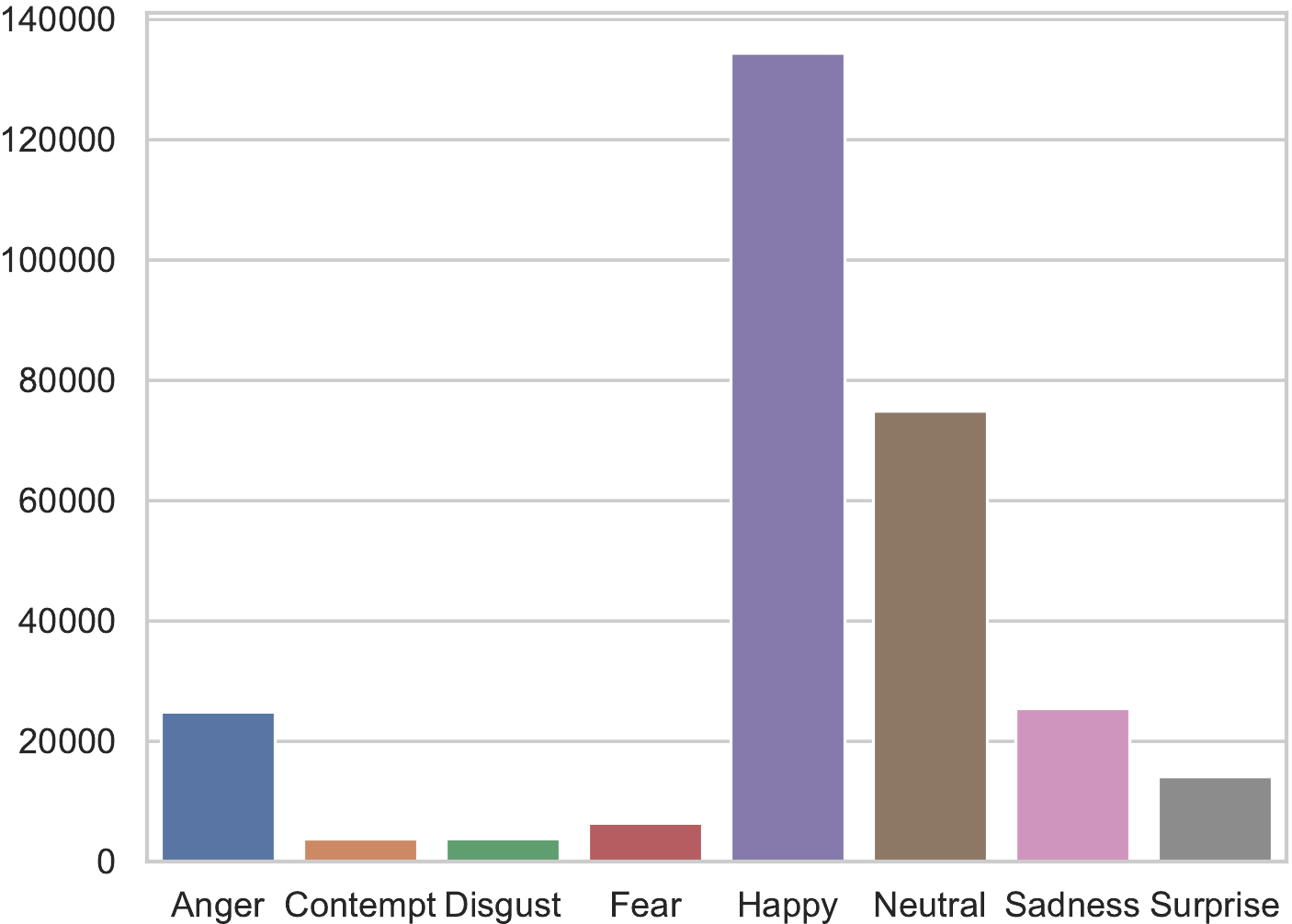}
    \caption{Distribution of annotated examples of expressions on AffectNet data set.}
    \label{fig:dist_affectnet}
\end{figure}

Applications of deep learning algorithms and frameworks in different 
computer vision tasks such as image classification~\cite{resnet_cvpr2016,krizhevsky2012imagenet}, 
face recognition~\cite{taigman2014deepface,schroff2015facenet,bhattarai2016cvpr}, face attribute classification~\cite{liu2015deep,kang2015face,hand2017attributes,kalayeh2017improving,bhattarai2016deep} are not new anymore.
Deep learning algorithms have proven to improve the performance of such applications substantially. However,
the bottleneck of training these algorithms is the need of large volumes of data and resources, and collecting such large volumes of data is expensive,
daunting and requires experts. Some of the tasks such as face recognition, attribute recognition \etc have to face 
another level of obstacle due to privacy issues. Fig.~\ref{fig:dist_affectnet}
shows the distribution of annotated data from AffectNet~\cite{mollahosseini2017affectnet}, which is one of the largest
annotated data sets for face attribute classification. We can clearly observe
that there are some categories that have an \emph{insufficient} volume of data 
to train a deep network optimally.

\begin{figure}[t]
\centering
\includegraphics[width=1\linewidth]{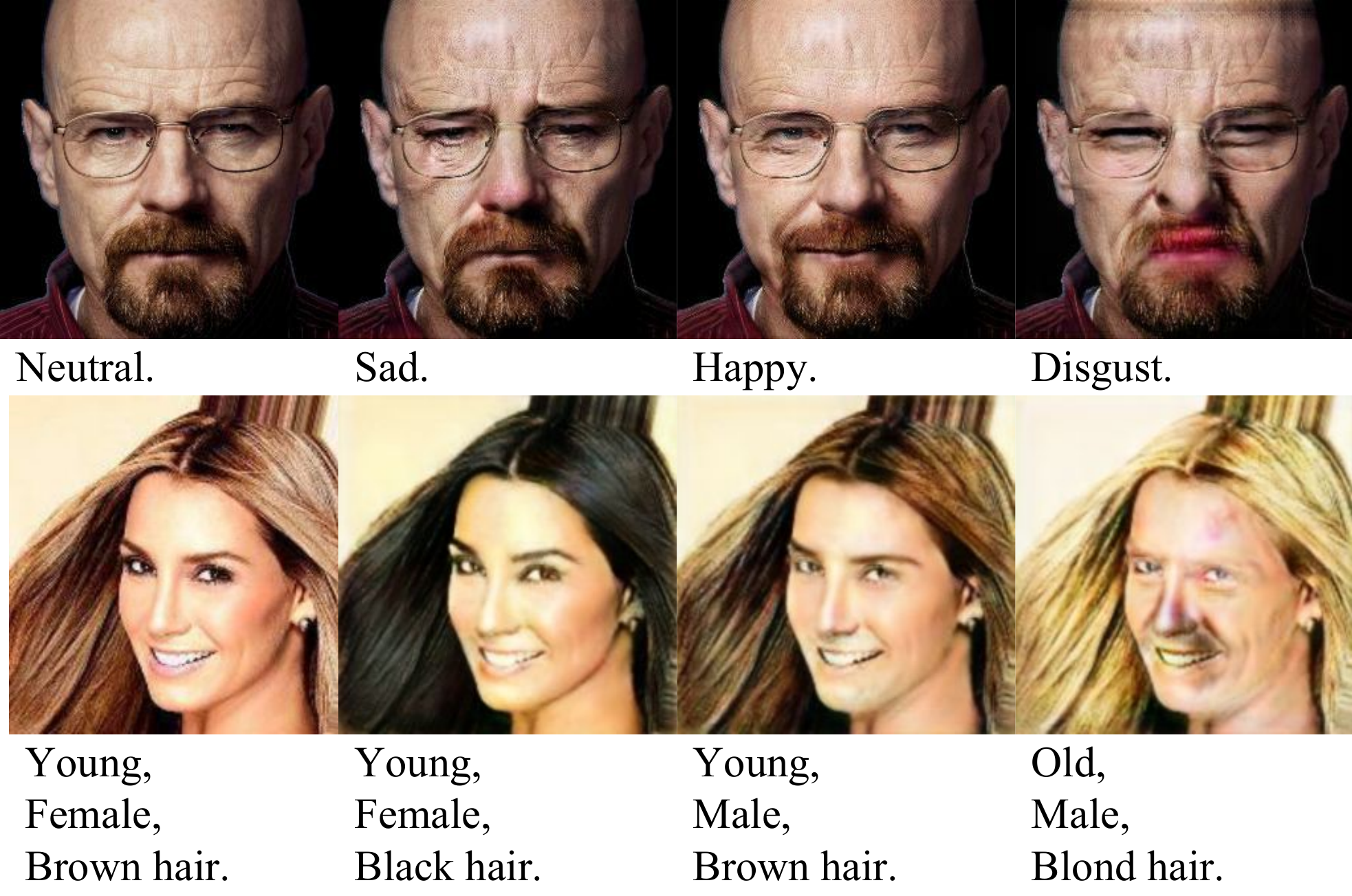}
\caption{Example emotion/attribute translation results generated by StarGAN~\cite{stargan_cvpr2018} on AffectNet dataset (Row 1) and CelebA dataset (Row 2), respectively. (Col. 1) Original image, (Col. 2-3) Rather successful translation, (Col. 4) Low quality translation. In this paper, we propose to filter out translated images having low-qualities. For this purpose, we propose three-types of simply implementable filters and empirically show their effectiveness.}
\vspace{-0.3cm}
\label{f:first}
\end{figure} 

To tackle such problems, research on augmenting the synthetic data with 
real data is growing these days~\cite{shrivastava2017learning,gecer2018eccv}. However, the research community is more focused on engineering the architecture of the deep networks in comparison to data engineering. There are several network architectures that are being proposed based on 
AlexNet~\cite{krizhevsky2012imagenet} to Inception Net~\cite{Googlenet_cvpr2015}, 
ResNet~\cite{resnet_cvpr2016}, a few to mention. 
In this paper, we propose methods to engineer the training data by discarding unwanted synthetic data before augmenting with real data.

One of the most common and successful methods to augment data to train a classification network is applying geometric transformations on 
images~\cite{krizhevsky2012imagenet}, such as rotation, translation,
flipping, cropping \etc. However, this technique does not guarantee that the label of the synthetic image will be preserved after applying such functions. Another study~\cite{hauberg2016dreaming} identifies 
the limitations of geometric transformation of not being able to 
preserve the label (\eg horizontal flip of 6 results into 9 in MNIST data set) of the
synthetic data in every case. Thus, feeding such examples during training hurts the performance of the model. To address this issue, \etal~\cite{cubuk2018autoaugment} recently proposed a
method to perform data specific geometric augmentation. Even then, methods of this category still depend on a single input image to generate multiple synthetic images.

Another line of research for data augmentation is the use of large synthetic data generated by GANs~\cite{dataaug_cvpr2018,shrivastava2017learning,zheng2017unlabeled,gecer2018eccv}. In these methods, synthetic data are used to augment real data 
but randomly when training CNNs.  Several GANs~\cite{stargan_cvpr2018,sagan_eccv18} are being proposed to generate synthetic examples by 
translating images from a source category to target categories.
Although the photo realism of the synthetic images generated by GANs is improving rapidly, even after augmenting millions 
of synthetic images, the improvement is still marginal. Recent study on \textit{Seeing is not necessarily believeling}~\cite{ravuri2019seeing}
observed that even after augmenting visually plausible synthetic examples the performance of the model is degraded. This 
could be due to large number of synthetic examples not preserving target label.  
Another study on power of GAN ~\cite{shmelkov2018good} demonstrates that
 random augmentation of synthetic images are not sufficient to improve the performance.
The inception score~\cite{barratt2018note} of images generated by most of the GANs are quite low (See Fig.~\ref{f:first}d). 
This entails, that most of the images do not preserve the target label and also lack realism.
Moreover, there is still a domain gap between real data and synthetic data. 
Some of the research works such as~\cite{shrivastava2017learning,realgap_cvpr2018} focus to minimize the domain gap between real and annotated
data. However, these methods rely on additional supervision
to align the parameters between real domain and target domain.
Due to these shortcomings on synthetic data, it is not useful to feed in all the synthetic examples to train CNNs. 

We are interested in mitigating the above mentioned challenges on synthetic data from GANs and maximize their benefits 
without using any external supervisions. Inspired from the success of seminal work on simple yet effective, sampling strategies of bag-of-features for image classification~\cite{nowak2006sampling}, we propose two different simple, effective and easy to implement 
approaches to sub-sample useful synthetic data from a large volume of synthetic data. 
Our methods are less demanding since we are mostly relying on the information,
which is available on the GAN itself and do not need additional annotations/source of information. One of them is target label preserving confidence score of synthetic examples, which is easy to compute from a pre-trained classifier on limited real examples. Another one is the confidence score of the realism of the synthetic data, which can be easily 
computed from the discriminator.
Finally, we propose to learn a policy to augment or not to augment the synthetic data using a reinforcement learning algorithm. Reinforcement Learning algorithms are successful for 
learning from experiences where there are no annotated examples available.

To validate our ideas we applied our method on two different challenging  face attribute classification data sets viz. CelebA~\cite{liu2015deep} 
and AffectNet~\cite{mollahosseini2017affectnet}.
We use StarGAN~\cite{stargan_cvpr2018}, which is one of the state-of-the-art face attribute translation GANs. 
We performed extensive experiments to validate our idea. 
To the best of our knowledge, this is the first work to do such systematic study on selecting the useful synthetic data from a pool of millions of 
synthetic data.

We summarize our contributions in the following points:
\begin{itemize}
    \item Two different efficient, effective and easy to implement data sampling methods
    \item Applied RL algorithm for sub-sampling GAN synthetic data
    \item Extensive systematic empirical experiments demonstrating the need of sub-sampling meaningful data.
    \item Improving the performance of state-of-the art deep architecture \textit{in vitro}.
\end{itemize}
\section{Related Works}
\label{related_work}
In this Sec., we further investigate related works that exploit large volumes of synthetic data to augment real data for training their models. Moreover, we will also present some of the works 
related to face attributes and expressions classifications.

\noindent \textbf{Geometric data augmentation.} Data augmentation has been getting popular after the use of CNNs to train a model. This is mainly due to the data voracious nature of CNNs. 
~\cite{krizhevsky2012imagenet} proposed geometric transformations (random flipping, random cropping \etc)  to generate synthetic examples to train their large scale CNN for image classification.
Similarly,~\cite{hauberg2016dreaming} proposed to learn data specific geometric transformations to train an image classification network. Recently~\cite{cubuk2018autoaugment} proposed image specific geometric transformation using reinforcement learning algorithms. This methods have improved the performance in comparison to their baselines trained on real data only.
~\cite{devries2017dataset} proposed to do geometric augmentation by adding noise on features, interpolating and extrapolating between features \etc.  Similarly, ~\cite{lemley2017smart}
proposed a network to automatically generate synthetic image by merging two or more samples from the same class.

\noindent \textbf{Synthetic model.}
Synthetic model suggests an easy way of collecting both 2D images and their corresponding labels, and they have also been used for collecting large-scale database~\cite{synth_cvpr2017,masi2016we}.
Compared to simple transformations used in data augmentation methods, synthetic models can supply quality data having diverse variations and semantics. However, the issue is the gap between real and synthetic data.  
Some of the recent works such as ~\cite{shrivastava2017learning} made attempts to tackle this gap problem.

\noindent \textbf{Generative Adversarial Networks (GANs).} The generative networks such as GANs and Variational Auto Encoders~(VAEs) can be used to generate new samples. Especially, GANs are known to be able to generate realistic samples, while the discriminator and the generator play a ``two-player minimax game''. Generating new type data using GANs and augementing with real data has been investigated in recent 
works~\cite{dataaug_cvpr2018,gecer2018eccv,sagan_eccv18,shmelkov2018good,zhao20183d,tran2017disentangled,zhao2018towards,huang2017beyond} and too few to mention. In this paper, we try to investigate methods and tricks to sub-sample instead of randomly augmenting the synthetic images from GAN. Please 
note that our methods are generic and can be applied for VAEs synthetic data too.

\noindent \textbf{Reinforcement Learning.} 
Recently, ~\cite{cubuk2018autoaugment} applied it for learning automatic polices to find the optimal geometric transformation to generate new examples.
However, we applied RL to sub-sample the GAN synthetic data. From our best knowledge, this is the first work to apply RL to sub-sample GAN synthetic data.
Another work on \textit{learn to simulate}~\cite{learnsimulate2019} applied RL to learn the optimal parameters of simulator to generate 
synthetic data. This method needs to be applied end to end fashion and hence, remains specific to a model. Our method can be applied to any of the pre-trained simulators.

\noindent\textbf{Face Attributes and Expression Classification}
Face attributes and expressions classification is one of the challenging and popular research problems. One of the seminal works on face attributes recognition is from Kumar \etal~\cite{kumar2009attribute}. They propose to 
learn a classifier for each of the attributes (\emph{hair, shape of nose, gender \etc}) and use the output of the classifier to encode faces for face verification purposes. ~\cite{hand2017attributes} propose to jointly learn 
the parameters of the face attributes which share common traits. Similarly,~\cite{rudd2016moon} proposed a joint optimisation function to model the parameters of different attributes together.~\cite{liu2015deep} proposes 
a cascade of deep networks to predict attributes of unaligned faces and also proposes CelebA, one of the largest databases for face attributes classification. In~\cite{zhang2014panda}, a pose-normalized CNN is proposed to estimate 
the attributes. Similarly, ~\cite{kalayeh2017improving} proposed to use semantic segmentation as privilege information to train a deep CNN. Recently, ~\cite{sun2018bmvc} proposed to learn hierarchical CNNs for attributes classification. Above all, most of the works are focused on designing the architecture of CNNs. However, our work is focused to design the training data set by selecting the useful ones. We suggest  readers to refer to the survey on expression classification~\cite{li2018deep} for more information.
\begin{figure*}
    \centering
    \includegraphics[trim= 0 0 0 0, clip, width=0.9\textwidth]{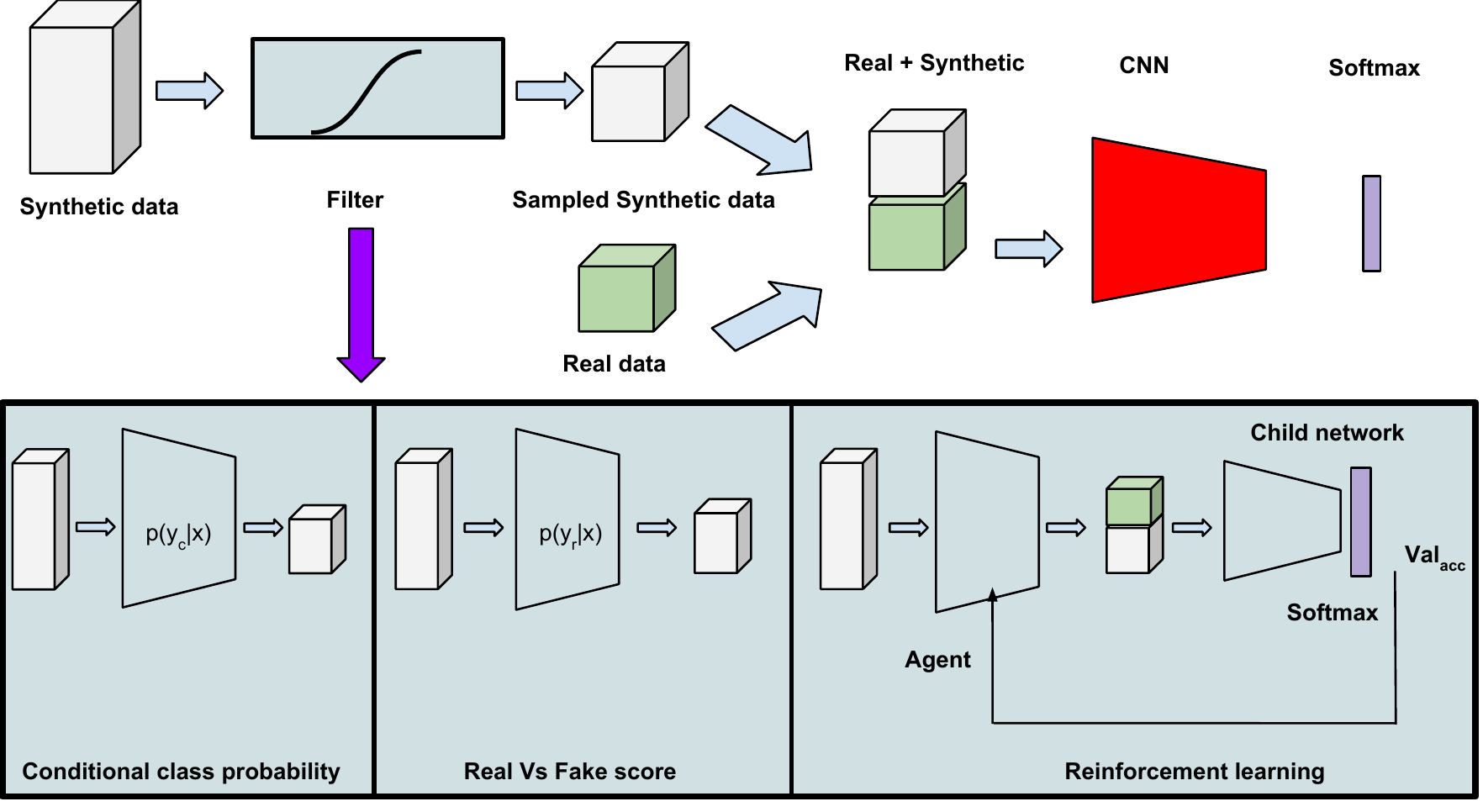}
    \caption{Schematic diagram of the proposed method. We propose to have three types of filters (\eg Conditional class probability(cl-sam), Real vs. Fake score (cr-sam) and Reinforcement learning(RL)) to get rid of unwanted synthetic data.}
    \label{fig:proposed_method}
\end{figure*}

\section{Proposed Method}
\label{methods}
In this section, we describe our proposed methods in a detailed level. Given real data $D_r=\{(x_i^r, y_i^r)\}_{i=1}^{i=L}, (x^r, y^r) \sim p_r(X^r, Y^r; \theta^r)$ and synthetic data $D_s=\{(x_i^s, y_i^s)\}_{i=1}^{i=M},  (x_s, y_s) \sim p_s(X_s, Y_s;
\theta_g)$ we are interested in sub-sampling the synthetic data. Here, $\theta_g$ is the parameter of the generator 
and $\theta_r$ is the distribution of the real data set, which is known only empirically. We have a 
scenario where $L<M$ and our objective is to select $N$ number of synthetic examples s.t. $N<<M$ and augment with the real data set $ \{(x_i^r, y_i^r)\}_{i=1}^{i=L}
\bigcup \{(x_i^s, y_i^s)\}_{i=1}^{i=N}$to train a model. It is important that we will improve the performance of the classifier on real validation data set.

\begin{figure*}
    \centering
    \includegraphics[trim=0.8cm 0 0 0,clip, width=0.33\linewidth]{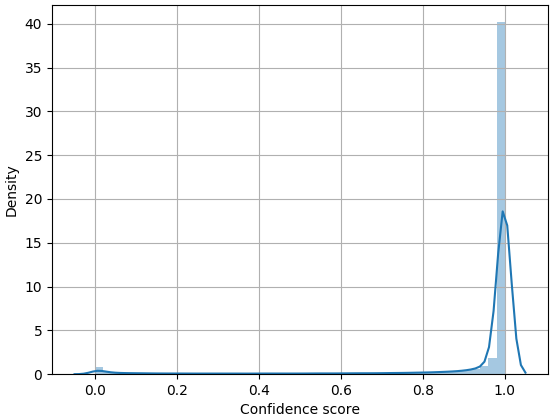}
    \includegraphics[trim=0.8cm 0 0 0,clip, width=0.33\linewidth]{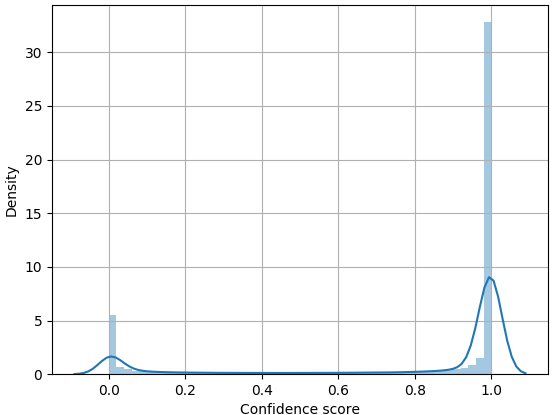}\\
    \caption{Distribution of confidence score on AffectNet synthetic data. X-axis represents the confidence score and Y-axis represents the distribution of Data (in \%). Order of the expressions are:
    Anger (left) and Contempt (right)}
    \label{fig:conf_density_affectnet}
\end{figure*}

To sub-sample the examples from synthetic data, we took two different approaches, which are similar to instance re-weighting for domain adaptation problem~\cite{jiang2007instance}. In our case, the instance weighting is in binary fashion : $1$ to select augmentation while $0$ for discarding the example. We set the threshold using two approaches from our prior knowledge. Without loss of generality, it is very important
for the synthetic data to be as realistic as possible and to preserve its target label \ie the class conditional probability should be high. However, in reality, there are many synthetic examples that do not preserve 
class conditional probability and also realism. Fig.~\ref{fig:conf_density_affectnet} shows the distribution of class-confidence score $p(y_s|x_s; \theta_c)$ ($\theta_c$ is the classifier model trained on real data set) 
predicted by the model trained on real data set. We can see that many synthetic examples from each category deviates from high confidence to low confidence. These examples will be misleading to train 
the model. 

Fig.~\ref{fig:proposed_method} shows the schematic diagram of our proposed pipelines. First, the data generated by the generator is passed through the data-sampler. There are three different types of data sampling techniques based on class 
conditional confidence score, realism conditional score and reinforcement learning. In this work, we are evaluating one sub-sampler at a time. The sub-sampler discards the unwanted data and lets only pass the useful data points. The filtered synthetic data is then augmented with the real data set and used to train the classifier. The volume of the data set, which is discarded is comparatively larger, in the order of few folds, in comparison to the passed data to train the final classifier. We elaborate on the size of the discarded volume of synthetic data in the supplementary sections. We discuss about the sub-sampling functions and the generator in the following sub-sections.

\noindent \textbf{Generator:}
We employed StarGAN~\cite{stargan_cvpr2018} as our generator. To reiterate, our methods are generic and can be used with any other types of GANs or
generators. StarGAN takes the source image, and target label as input and returns the translated image. In the similar way, it also takes the synthetic
image and source label as input and reconstructs the original image. For attributes synthesis, we used the publicly available pre-trained
model, whereas for expression synthesis, we used training data from AffectNet, one of the largest data sets
annotated with different expressions. For expression synthesis, we trained the model from scratch. 

\noindent\textbf{Class conditional probability (cl-sam):}
We propose to use class conditional probability, which is commonly known as class confidence score, as one of the filters to discard the unwanted examples. For a given synthetic example, we computed class 
conditional probability $P(y_c|x_s; \theta_c)$. Here, $x_s$ represents the synthetic data, $y_c$ the target class $c$ and $\theta_c$ the model parameters of the classifier trained on real data only. 
This confidence score is utilized to filter out the synthetic examples. We rank the synthetic examples based on the conditional target class label(on descending order) and select the top-$K$. For the parameters of the $\theta_c$, we employ ResNet50 architecture~\cite{resnet_cvpr2016} and train it using the stochastic gradient descent (SGD) by minimizing categorical cross-entropy loss:
\begin{equation}
\label{loss:ce}
\mathcal{L}  = -\frac{1}{N}\sum_{i=1}^{i=N}\sum_{j=1}^{j=C} y_{t, i}^j\log y_{p, i}^j 
\end{equation}
Once the classifier is trained, we used this classifier to score the synthetic data and rank them in descending order. We then selected the top-$K$ of the
synthetic images from each category and augmented with the real data set.
We called the sampler based on this score as \emph{cl-sam}.

\noindent\textbf{Realism conditional probability (rl-sam):}
It is equally important that the synthetic examples are as realistic as possible.  
We propose to use the confidence on realism as another parameter for our sampling function. We use parameters of the discriminator to compute the realism confidence score on synthetic data. We then rank them (in descending order) for each category. 
The top-$K$ are selected to augment the real training data set. Similar to the previous one, we train our classifier again with these new training examples.
We called the sampler based on this score as \emph{rl-sam}.

\noindent\textbf{Reinforcement Learning:}
We explore using reinforcement learning setup to 
select the synthetic data, which makes the model more discriminative, and reduces the redundancies and unwanted noisy data. 

We choose a subset of the real training data ~1\% and select $8\times$ large synthetic data. 
We assume a scenario where synthetic data is abundant and real data is limited. We train the policy network 
of 3 (a CNN with 3 convolutional layers and 
2 fully connected layers) to sub-sample the synthetic examples. Our policy network takes image as input, thus the
policies are conditioned on the content of the images (this is the main difference from ~\cite{cubuk2018autoaugment}).
Fig.~\ref{fig:proposed_method} shows the schematic diagram of the proposed method. We use the actor and critic
method similar to~\cite{cubuk2018autoaugment} to learn the augmentation policies.

\noindent \textbf{Reward.}
We compute the reward based on the score on validation set using the child network, similar to ~\cite{cubuk2018autoaugment}.
The child network is a small classification network, which mimics the final classification network. The architecture 
is set same as the aforementioned policy network. We compare the val score with the threshold score. We compute 
threshold by averaging the val scores in the sliding window of last 5 episodes. If the score is higher than the threshold, 
we assign +1 to policy otherwise -1.

\section{Experiments}
\label{experiments}
In this section, we give information about the dataset we use, the pre-processing method, evaluation protocol and compared baselines. Then, we analyze our experimental results.

\subsection{Datasets}

\noindent \textbf{CelebA.}
This is one of the largest and most widely used datasets for attribute classification. This data-set consists of 200K annotated examples and is divided into training, validation and testing set of sizes of 160K, 20K and 20K, respectively. There are 40 attributions in total. For our experiments, we have selected 5 important attributes.  

\noindent \textbf{AffectNet.}
This dataset is one of the largest datasets for expression, emotions and valence arousal estimations. 
In this dataset, there are nearly 1M samples where ~400K of them are manually annotated and the rest is automatically
annotated. The manually annotated dataset is further divided into up-sampled and down-sampled cases to
handle the imbalance number of annotations of different expressions. We choose the sub-sampled one for our
evaluation purpose. This version consists of ~88K annotated examples. The images are annotated with 8
different expressions and split into train, val and test set. We use these 8 expression annotations for our training and evaluation purpose.

 \noindent \textbf{Synthetic data.}
 We use StarGAN~\cite{stargan_cvpr2018}, one of state-of-the-arts GANs, to generate the synthetic examples. Note that any kind of GANs could be used here. For CelebA,
 we use publicly available pre-trained models, whereas for AffectNet, we use the training set to train 
 StarGAN from  scratch. We generated synthetic data up to 12-folds and 7-folds of real data for CelebA and AffectNet, respectively.  
 
\subsection{Preprocessing of the data}
We train the CNN at different resolutions. Tab~\ref{tab:exp_affectnet} shows the 
baseline performance on AffectNet at different resolutions. As we observe that
$128\times128\times3$ attains the performance of previous method
reported on ~\cite{mollahosseini2017affectnet} with the resolution of $224\times224\times3$, we set this resolution for 
further evaluations. For CelebA, we choose $64 \times 64 \times 3$  to reduce the computing complexity,
since this dataset is comparatively larger. We resize CelebA  and AffectNet to $72 \times 72 \times 3$ 
and $144 \times 144 \times 3$, respectively and randomly crop on 4 corners and centre. We also randomly flip the images when training the network, while at test time, we centre crop the images.

\subsection{Evaluation protocol}
We compute attribute classification accuracies on two benchmark datasets: CelebA and AffectNet dataset to
evaluate the proposed methods for quantitative evaluations. We also provide qualitative visualisations to compare the quality of the images sampled by the evaluated methods.

\subsection{Evaluated methods}
We have experimented with $5$ baseline methods using the state-of-the-art ResNet-50 architecture.

\noindent \textbf{Real data set.} This is the most commonly used and successful data augmentation technique to this date.
One of the baselines for us is the deep CNN trained on large scale dataset. As we
know, CNNs are trained on very large scale data and the performance is near saturation.

\noindent \textbf{Random augmentation.} We randomly sub-sample the synthetic set on different proportions ($1\times,2\times, 5\times$) compared to real data. We augment this data with real data to re-train the CNNs from scracth. In case of AffectNet, we initialized the network with weights of a pre-trained network for ImageNet for all cases.

\noindent \textbf{Conditional class conf. sampler (cl-sam)} We compute the class confidence scores on synthetic data by the model trained on real dataset only. As we mentioned before, we rank them based on the confidence score in descending order and select top-K (where 
 $K=1\times, 2\times, 5\times$) of the synthetic data. We then augment this synthetic data to train the CNN
from the beginning.

\noindent \textbf{Discriminator Real/Fake score-based sampler (cr-sam).} For each synthetic example translated to target category, we
computed the discriminator's real vs fake  score. Similar to the confidence score set-up, we re-ranked 
the synthetic data and selected the top-K examples to augment the real dataset.

\noindent \textbf{Reinforcement learning (RL)-based sampler.} As we discussed before, we train the agent to select the useful synthetic data. This agent is applied to the whole synthetic data and only augmented the synthetic data chosen by the agent. 

\begin{table*}
    \centering
    \resizebox{1.0\textwidth}{!}{
    \begin{tabular}{c|c|c|c|c|c|c|c|c|c|c}
     Architecture & Resolution & Black Hair & Brown Hair & Blond Hair & Female vs. Male & Young vs. Old & Mean. Acc. & Aug. & Type \\
     \hline
     \cite{celeb_comp1} & $224\times224\times3$ & $70$ & $60$ & $80$ & $91$ & $80$ & $80.1$ & $0\times$ &No aug.\\
     \hline
     \cite{liu2015deep} & $224\times224\times3$ & $88$ & $80$ & $95$ & $98$ & $87$ & $87.3$ & $0\times$ &No aug.\\
     \hline
     \cite{rudd2016moon} & $224\times224\times3$ & $89.4$ & $89.4$ & $95.9$ & $98.1$ & $88.1$ & $90.9$ & $0\times$ &No aug.\\
     \hline
     \cite{celeb_comp4} & $224\times224\times3$ & $84$ & $81$ & $92$ & $96$ & $86$ & $88.7$ & $0\times$ &No aug.\\
     \hline
     \cite{celeb_comp5} & $224\times224\times3$ & $90.5$ & $88.5$ & $96.2$ & $98.2$ & $88.9$ & $91.5$ & $0\times$ &No aug.\\
     \hline
     \cite{kalayeh2017improving} & $224\times224\times3$ & $90.1$ & $89.2$ & $95.8$ & $97.7$ & $87.8$ & $91.2$ & $0\times$ &No aug.\\
     \hline
     \cite{sun2018bmvc} & $224\times224\times3$ & $90.2$ & $89.0$ & $96.1$ & $98.8$ & $88.9$ & $91.6$ & $0\times$ &No aug.\\
      \hline
     \hline
     ResNet-50 & $64\times 64\times 3$ & $87.8$ & $86.2$ & $95.0$ & $97.2$ & $85.4$ & $90.3$ & $0\times$ & No aug. \\
     \hline
     \hline
     ResNet-50 & $64\times 64\times 3$ & $88.7$ & $87.3$ & $95.3$ & $97.3$ & $86.6$ & $91.0$ & $5\times$ & Random \\
     \hline
     ResNet-50 & $64\times 64\times 3$ & $88.9$ & $87.1$ & $95.5$ & $96.8$ & $87.1$ & $91.1$ & $5\times$ & cl-sam. \\
     \hline
     ResNet-50 & $64\times 64\times 3$ & $88.7$ & $87.4$ & $95.4$ & $97.2$ & $86.8$ & $91.0$ & $5\times$ & cr-sam. \\
     \hline
    \end{tabular}
   }
    \caption{Comparison of mean average performance our evaluated methods with existing art on CelebA.}
    \label{tab:perf_baseline_celeba}
\end{table*}

\subsection{Experimental results}
We performed extensive evaluations on proposed methods to validate the ideas. 
We have compared our methods on two challenging face attributes classification and expression classifications data set. 
In the following sub-section, we analyse our results in detail. We will first start with quantitative analysis followed by qualitative analysis.

\noindent \textbf{Quantitative analysis} \\
\noindent \textbf{Baseline} Tab.\ref{tab:exp_affectnet} shows the mean accuracies of the compared methods and existing art on AffectNet.

From Tab. we can observe that the performance of our baseline implementation 
ResNet-50 on $(128\times128\times3)$ is slightly lower than existing art ($-0.4\%$). We set this architecture and resolution as our baseline and performed further analysis. 

\noindent \textbf{Augmentation} We then augmented real data with different proportions sampled by the evaluated methods. As we expected, the performance on the test set improves as the augmentation size is increased from $0\times$ to $1\times$ in all cases. However, we observe a difference in
performance gain between the evaluated methods. Random augmentation yields the minimum gain ($+0.4\%$) whilst \emph{rl-sam} (based on realism) yields the highest gain ($+2.6\%$). Similarly, \emph{cl-sam} observed the gain of ($+2.1\%$). This is expected, 
as the random method samples both useful and and misleading examples, while \emph{rl-sam} and \emph{cl-sam} manage to collect examples that are more realistic and preserve the class-conditional label, respectively.  
On further increasing the volume of synthetic data we observe further improvement on the performance of random and \emph{cl-sam}, while the performance of \emph{cr-sam} is slightly degraded. It is because being real does not ensure target 
label of the synthetic data is preserved. The ratio of the performance improvement from $1\times$ to $2\times$ augmentation
was lower than when augmentation is of size $1\times$. On further increasing the augmentation size to $5\times$, we observe degradation of the performance of all the three methods in comparison to $2\times$ augmentation. However, the performance of cl-sam is degraded by a
minimum margin while the degradation of performance by random sampling is maximum. This supports the fact that there are only a limited number of useful data to augment. With increase in size of augmentation, the ratio of useful synthetic data to misleading data keeps on decreasing. This trend of performance is further supported by  Fig.~\ref{fig:conf_density_affectnet}. In the Fig., we can clearly see a large number of misleading examples \ie with low class confidence score. There is similar trend on realism score too. Please refer supplementary material for more details.
In addition to this, we also applied our RL policy to sub-sample the synthetic data. It selected only $2.6\times$ of real data of synthetic data ($7\times$ of real data is the size of full synthetic data).
The performance of the RL in comparison to cl-sam is slightly lower. However, it outperforms the performance of the other two methods. As we know, \emph{cl-sam} was trained with a real training set of size ~$88$K 
data to learn the parameters, whereas RL uses no such annotations but learns only from experience. Another potential reason for RL not being as competitive as \emph{cl-sam} is due to huge difference in architecture of 
child network and final classification network. For us, child network has comparatively very less parameters and different architecture. As we mentioned before, our child network has 3 Conv layers and 2 fully connected layers. 
Whereas, our classification network is Resnet50. Thus, the policies learned for child network may not be necessarily generalisable to large classification network. It will be computationally highly expensive to have a child network with 
the parameters similar to that of Resent50. 

We also compared category level of expression accuracy of the compared methods. Tab.~\ref{f:affectnet_mainexp} shows the categorical performance comparison between all the compared methods on Affectnet. We can see that random augmentation suffers in
wide range of performance gain and drop. For example, \emph{Contempt} improves from $72.2\%$ to $90.2\%$ when the augmentation size is increased from $1\times$ to $5\times$. In the similar range, the performance of  \emph{Sadnees}
drops from $60.9\%$ to $46.5\%$. We did not observe such trends on other compared methods. This suggests that the model trained with randomly augmented data are less robust to other approaches. 

Similarly, we also performed extensive experiments on CelebA, another challenging and widely used data sets for the attributes classification. We observe the similar trends that we observe on Affectnet. Please refer Tab.~\ref{tab:perf_baseline_celeba}
for more details. As We observed on Affectnet, \emph{cl-sam} is outperforming other compared method. We also compared our performance with several state-of-the-arts method. Even though we performed our experiments on $4\times$ lower resolution 
\ie $(64\times64\times3)$  than compared arts, our methods are either outperforming or competitive. Similarly, Tab.~\ref{f:celeba_mainexp} shows the categorical attribute classification performance on CelebA.

\begin{figure*}
\centering
\includegraphics[trim= 0 1.1cm 0 0, clip,width=1.0\linewidth, height=0.30\linewidth]{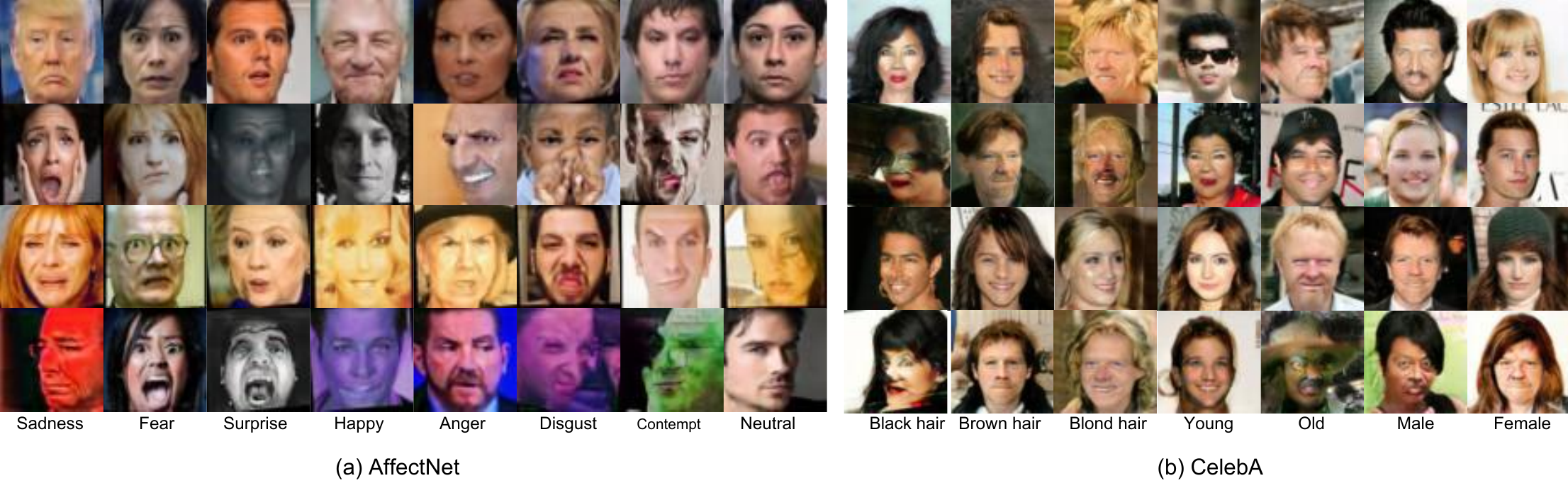}
\vspace{-0.3cm}
\caption{Example visualization according to the filter scores on AffectNet(left) and CelebA (right). Each row shows samples with different scores: (Row 1) Samples with {\color{red}High}  \emph{Class Confidence} score, (Row 2) Samples with {\color{blue}Low} \emph{Class Confidence} score, (Row 3) Samples with {\color{red}High} \emph{Real vs. Fake} score, (Row 4) Samples with {\color{blue}Low} \emph{Real vs. Fake} score. Each column represents samples from different categories. (Best viewed in color)}
\label{f:qual}
\end{figure*}

\begin{figure}
    \centering
    \includegraphics[width=0.45\textwidth]{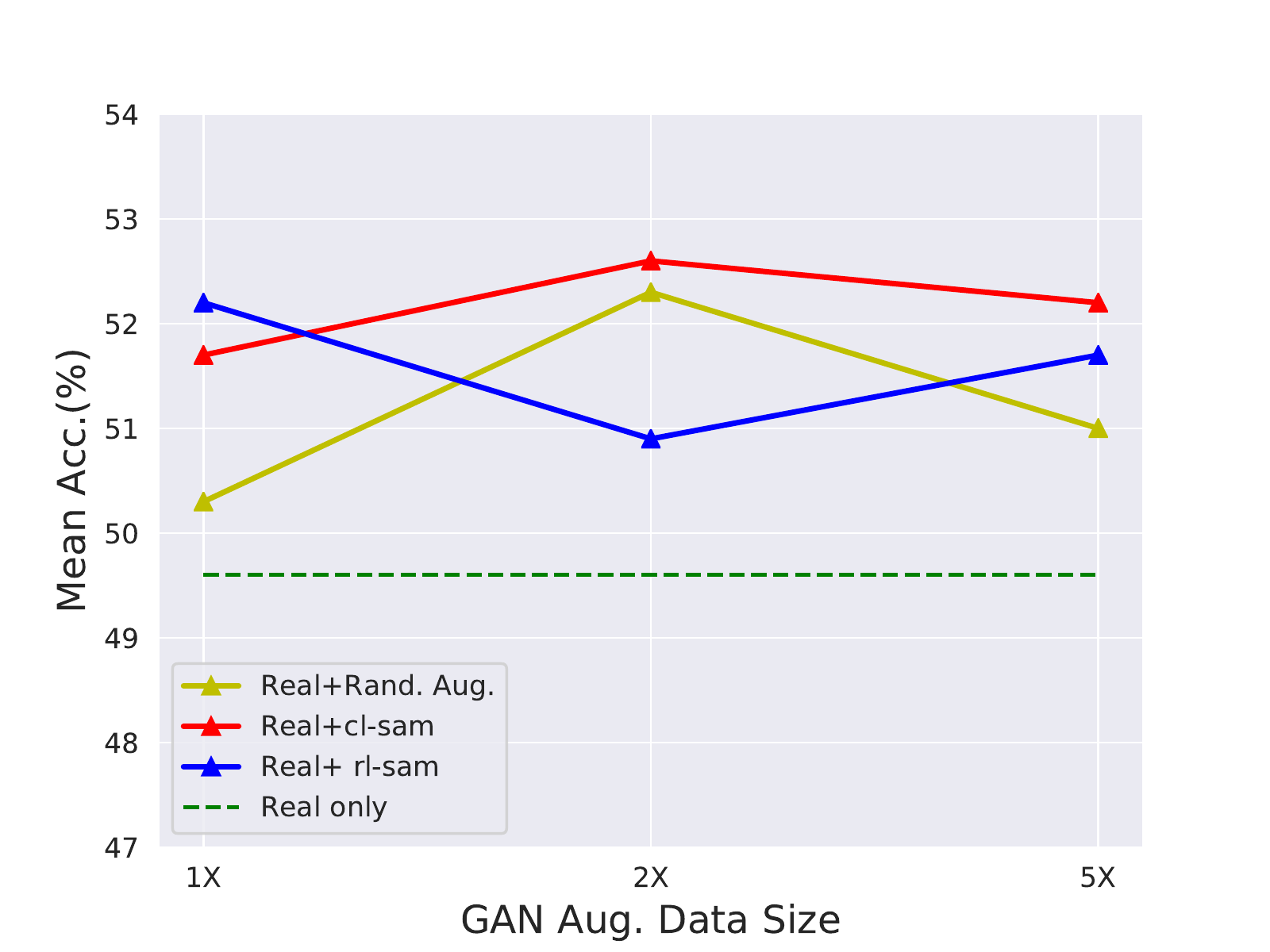}
    \caption{Mean performance comparison of different sampling strategies on AffectNet.}
    \label{fig:mean_perf_comp_affectnet}
\end{figure}

\noindent \textbf{Qualitative analysis.} 
In this sub-section we present our qualitative evaluations. In Fig.~\ref{f:qual}, left block shows the synthetic images from Affectnet while right block shows the CelebA synthetic images. 
In each block, each column shows a sample from each of the categories of expressions (Affectnet) and attributes (CelebA).  While each row shows (from top to bottom) the samples having {\color{red}high}/{\color{blue}low} 
\emph{Class conditional probability} and \emph{Real vs. Fake score}.
We can clearly observe that images with high confidence scores are more visually plausible and appealing than low confidence scores. Similarly, images with high realism scores are better in quality than 
lower realism scores. This further supports our argument and also our empirical evidences.

\begin{table}[]
    \centering
    \resizebox{0.5\textwidth}{!}{
    \begin{tabular}{c|c|c|c|c}
     Architecture & Resolution & Mean. Acc. & Aug. & Type \\
     \hline
     AlexNet~\cite{mollahosseini2017affectnet} & $224\times224\times3$ & $\textbf{50.0}$ & $0\times$ &No aug.\\
     \hline
     ResNet-50 & $64\times 64\times 3$ & $46.1$ & $0\times$ & No aug. \\
     \hline
     ResNet-50 & $128\times 128\times 3$ & $49.6$ & $0\times$ & No aug. \\
     \hline
     \hline
     ResNet-50 & $128\times 128\times 3$ & $50.3$ & $1\times$ & Random \\
     \hline
     ResNet-50 & $128\times 128\times 3$ & $51.7$ & $1\times$ & cl-sam \\
     \hline
     ResNet-50 & $128\times 128\times 3$ & $\textbf{52.2}$ & $1\times$ & cr-sam \\
     \hline
     \hline
     ResNet-50 & $128\times 128\times 3$ & $52.3 $ & $2\times$ & Random \\
     \hline
     ResNet-50 & $128\times 128\times 3$ & $\textbf{52.6} $ & $2\times$ & cl-sam \\
     \hline
     ResNet-50 & $128\times 128\times 3$ & $50.9$ & $2\times$ & cr-sam \\
     \hline
     \hline
     ResNet-50 & $128\times 128\times 3$ & $51.0 $ & $5\times$ & Random \\
     \hline
     ResNet-50 & $128\times 128\times 3$ & $\textbf{52.2} $ & $5\times$ & cl-sam \\
     \hline
     ResNet-50 & $128\times 128\times 3$ & $51.7$ & $5\times$ & cr-sam \\
     \hline
     \hline
     ResNet-50 & $128\times 128\times 3$ & $51.8$ & $2.6\times$ & RL \\
     \hline
    \end{tabular}
    }
    \caption{Comparison of mean average performance our evaluated methods with existing art on AffectNet.}
    \label{tab:exp_affectnet}
\end{table}

\begin{table}[]
\resizebox{0.5\textwidth}{!}{
\centering
 \begin{tabular}{cc|c|c|c|c|c|c|c|c|c|}
\cline{3-10}
& & \multicolumn{8}{ c| }{Expressions} \\ \cline{3-10}
& & Anger & Contempt & Disgust & Fear & Happy & Neutral & Sadness & Surprise \\ \cline{1-10}

\multicolumn{1}{ |c| }{\multirow{1}{*}{$0\times$} } &
\multicolumn{1}{ |c| }{Real} & 41.4 & 62.5 & 64.5  & 68.7 & 55.0 & 39.2 & 50.6 & 44.0 \\ \cline{1-10}
\hline 
\hline 

\multicolumn{1}{ |c| }{\multirow{3}{*}{$1\times$} } &
\multicolumn{1}{ |c| }{Random} & 36.6  & \textbf{72.2} & 69.3 &  \textbf{77.3} & \textbf{60.9}& 36.7 & \textbf{60.9} & \text{51.6} \\ \cline{2-10}
\multicolumn{1}{ |c  }{}                        &
\multicolumn{1}{ |c| }{cl-sam} & 46.6  & 60.4  & 68.9 &  67.3 & 55.2  & 40.0 & 46.8 & 50.2 \\ \cline{2-10}
\multicolumn{1}{ |c  }{}                        &
\multicolumn{1}{ |c| }{cr-sam} & \textbf{47.2} &  68.8 & \textbf{77.8} & 66.3 & 55.7 & \textbf{40.2}  & 52.7 & 47.8 \\ \cline{1-10}
\hline
\hline
\multicolumn{1}{ |c  }{\multirow{3}{*}{$2\times$} } &
\multicolumn{1}{ |c| }{Random} & 46.5 &  68.4 & 66.2  & 60.7  & \textbf{62.5}  & 39.9 & 50.2 &  46.6 \\ \cline{2-10}
\multicolumn{1}{ |c  }{}                        &
\multicolumn{1}{ |c| }{cl-sam} & \textbf{48.9} & 71.5 & 75.6 & 71.6 & 56.3 &  \textbf{40.5} &  51.0 &  \textbf{48.9} \\ \cline{2-10}
\multicolumn{1}{ |c  }{}                        &
\multicolumn{1}{ |c| }{cr-sam} & 47.3 & \textbf{74.8} & \textbf{78.9}  &  \textbf{71.7} & 55.7 & 36.1 & \textbf{65.4} & 43.8 \\ \cline{1-10}
\hline
\hline 
\multicolumn{1}{ |c  }{\multirow{2}{*}{$5\times$} } &
\multicolumn{1}{ |c| }{Random} & 44.7 & \textbf{90.2} & \textbf{74.4} & 69.0  & 62.4 & \textbf{39.6} & 46.5 & 45.3 \\ \cline{2-10}
\multicolumn{1}{ |c  }{}                        &
\multicolumn{1}{ |c| }{cl-sam} & \textbf{51.0} & 63.6  & 68.4  &  \textbf{70.8} & 53.6 & 38.4 & 46.5 & \textbf{51.6} \\ \cline{2-10}
\multicolumn{1}{ |c  }{}                        &
\multicolumn{1}{ |c| }{cr-sam} & 49.0 &  64.3  &  67.5 & 67.5 & \textbf{62.9} & 35.4 & \textbf{49.5} & 46.4 \\ \cline{1-10}
\hline 
\multicolumn{1}{ |c  }{\multirow{1}{*}{2.6x} } &
\multicolumn{1}{ |c| }{RL} & 44.0 & 71.2  & 76.2 & 68.8 & 60.5 & 37.0 & 56.7 & 48.6 \\ \cline{1-10}
\end{tabular}
}
\caption{Comparison of categorical performances between our evaluated methods on AffectNet.}
\label{f:affectnet_mainexp}
\end{table}

\begin{table} 
\resizebox{0.5\textwidth}{!}{
\centering
 \begin{tabular}{cc|c|c|c|c|c||c|}
\cline{3-8}
& & \multicolumn{6}{ c| }{Attributes} \\ \cline{3-8}
& & Black Hair & Blonde Hair & Brown Hair & Female/Male & Young/Old & Micro Avg. \\ \cline{1-8}

\multicolumn{1}{ |c| }{\multirow{1}{*}{$0\times$} } &
\multicolumn{1}{ |c| }{Real} & 87.8 & 95.0 & 86.2 & 97.2 & 85.4 & 90.3 \\ \cline{1-8}
\hline 
\hline 

\multicolumn{1}{ |c| }{\multirow{3}{*}{$1\times$} } &
\multicolumn{1}{ |c| }{Random} & 88.2 & 95.0 & \textbf{87.1} & 97.2 & 86.3 & 90.8  \\ \cline{2-8}
\multicolumn{1}{ |c  }{}                        &
\multicolumn{1}{ |c| }{cl-sam} & 88.3 & 95.2 & 86.5 & \textbf{97.2} & \textbf{86.4} & 90.8  \\ \cline{2-8}
\multicolumn{1}{ |c  }{}                        &
\multicolumn{1}{ |c| }{cr-sam} & \textbf{88.5} & \textbf{95.4} & 87.0 & 97.1 & 85.9 & \textbf{90.8} \\ \cline{1-8}
\hline
\hline
\multicolumn{1}{ |c  }{\multirow{3}{*}{$2\times$} } &
\multicolumn{1}{ |c| }{Random} & 88.2 & 94.9  & \textbf{87.1} & \textbf{97.2} & 86.3  & 90.8  \\ \cline{2-8}
\multicolumn{1}{ |c  }{}                        &
\multicolumn{1}{ |c| }{cl-sam} & \textbf{89.0} & \textbf{95.7} & 86.5 & 97.1 & 86.2 & \textbf{90.9}  \\ \cline{2-8}
\multicolumn{1}{ |c  }{}                        &
\multicolumn{1}{ |c| }{cr-sam} & 88.9 & 95.4 & 86.7 & 96.8 & \textbf{86.8} & 90.8 \\ \cline{1-8}
\hline
\hline 
\multicolumn{1}{ |c  }{\multirow{2}{*}{$5\times$} } &
\multicolumn{1}{ |c| }{Random} & 88.7 & 95.3 & 87.3 & \textbf{97.3} & 86.6 & 91.0  \\ \cline{2-8}
\multicolumn{1}{ |c  }{}                        &
\multicolumn{1}{ |c| }{cl-sam} & \textbf{88.9} & \textbf{95.5} & 87.1 & 96.8 & \textbf{87.1} & \textbf{91.1}  \\ \cline{2-8}
\multicolumn{1}{ |c  }{}                        &
\multicolumn{1}{ |c| }{cr-sam} & 88.7 & 95.4 & \textbf{87.4} & 97.2 & 86.8 & 91.0  \\ \cline{1-8}

\hline 

\end{tabular}
}
\caption{Categorical expression performance comparison of the compared augmentation techniques for CelebA.}
\label{f:celeba_mainexp}
\end{table}

\section{Conclusions}
\label{conclusions}
In this paper we evaluated three different data augmentation techniques over random augmentation technique. Firstly, we propose to use confidence score based sampler to find a meaningful sub-set.
Similarly, we proposed to use realism conditional probability based sampler. Finally, we explored reinforcement learning based sampler, which learns from the experiences. 
From our extensive experiments, we observed that these three techniques outperform the commonly used random augmentation technique and improves the performance of state-of-the-art CNNs. Among these three, 
we observed that the class conditional based sampler performs the best followed by RL and realism conditional probability based sampler. Each method has its own shortcomings and advantages. Confidence scored based sampler requires 
real training examples. Although realism conditional based sampler does not require labelled training example, it does not guarantee to preserve the class conditional probability.
RL does not require training examples but it is computationally expensive.

\section{Acknowledgements}
This work is partly supported by EPSRC FACER2VM project.

\bibliography{egbib}
\bibliographystyle{aaai}
\end{document}